\documentclass[10pt,twocolumn,letterpaper]{article}

\usepackage[pagenumbers]{cvpr} 

\usepackage{xcolor}
\usepackage{flafter}

\definecolor{WiCompassTableGray}{RGB}{82,82,82}
\newcommand{\wbt}[1]{#1}
\newcommand{\wct}[1]{\textcolor{WiCompassTableGray}{#1}}

\definecolor{cvprblue}{rgb}{0.21,0.49,0.74}
\usepackage[pagebackref,breaklinks,colorlinks,allcolors=cvprblue]{hyperref}

\def\paperID{*****}
\def\confName{CVPR}
\def\confYear{2026}

\title{Wave2Body: Rethinking mmWave Human Pose Estimation \\as Radar-to-Body Token Translation}

\author{
Bo Liang\textsuperscript{1}
\quad
Chen Gong\textsuperscript{1}
\quad
Wei Gao\textsuperscript{2}
\quad
Chenren Xu\textsuperscript{1,3,*}
\\[2mm]
{\small \textsuperscript{1}School of Computer Science, Peking University} {\small \textsuperscript{2}University of Pittsburgh}
\\
{\small \textsuperscript{3}Key Laboratory of High Confidence Software Technologies, Ministry of Education (PKU)}
\\
{\small \textsuperscript{*}Corresponding author: chenren@pku.edu.cn}
}

\begin{document}

\maketitle

\begin{abstract}
Millimeter-wave (mmWave) radar enables privacy-friendly human sensing, but its sparse point clouds are physical measurements of view-dependent electromagnetic reflections and only indirectly characterize body articulation.
Recovering a complete 3D pose from such partial, geometry-dependent observations is therefore under-constrained. Existing methods directly regress joint coordinates from paired radar--pose data, relying on the same limited paired supervision to learn radar perception, human-body structure, and their alignment. This coupling can encourage dataset-specific shortcuts under ambiguous radar observations. We propose Wave2Body, a radar-to-body token translation framework that decouples these learning targets using a self-supervised mmWave tokenizer, a pretrained compositional body tokenizer that defines the output space, and a lightweight translator between them. Experiments on M4Human and mmBody show that Wave2Body achieves stronger cross-domain generalization than previous methods while incurring much lower computational costs for training and inference. All the code and experiment results are publicly available at \\
\textcolor{blue}{\texttt{https://github.com/Galaxywalk/Wave2Body}}.
\end{abstract}

\section{Introduction}
\label{sec:introduction}

\begin{figure}[t]
\centering
\captionsetup{skip=1pt}
\includegraphics[width=\columnwidth]{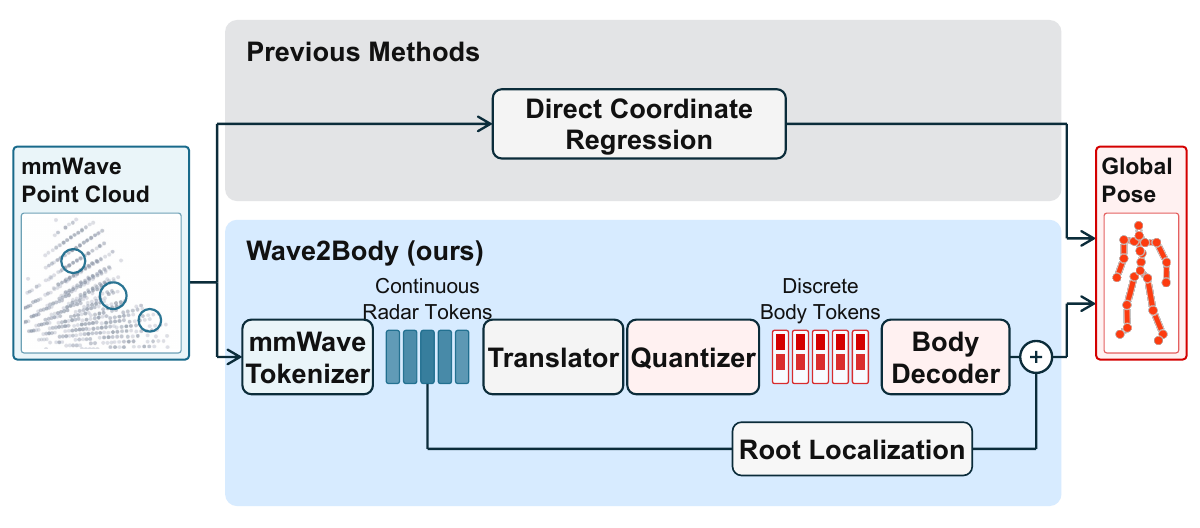}
\caption{\textbf{Wave2Body reformulates mmWave HPE from coordinate regression to radar-to-body token translation.} A self-supervised mmWave tokenizer produces radar tokens, which are translated and quantized into discrete body tokens before decoding; the body encoder is used only to construct training targets.}
\label{fig:wave2body_overview}
\vspace{-16pt}
\end{figure}

Millimeter-wave (mmWave) radar has emerged as a promising sensing modality for human motion understanding in mobile and ubiquitous systems. Unlike cameras, mmWave radar does not record visible appearance, can operate under poor illumination, and remains robust to partial visual occlusions. By reconstructing sparse 3D point clouds from radio reflections, recent mmWave-based human pose estimation (HPE) systems have enabled privacy-friendly 3D body perception for in-home monitoring, clinical rehabilitation, and human--robot interaction~\cite{gu2024millimeter, hu2024mmwave, hu2026waveman, zhang2023lt}. Progress~\cite{mmGPE,FUSE,mmdiff,mmJoints} has been made through stronger point-cloud backbones, signal augmentation, diffusion-based pose estimation, and reliability-aware modeling for pose recovery.

Despite this progress, mmWave-based HPE differs from vision-based HPE. Radar returns depend on body geometry, orientation, motion, signal propagation, and sensor configuration; they do not provide dense observations of anatomical joints. Some body parts may produce weak or missing returns, while reflections can admit multiple plausible body configurations. Recovering a complete 3D skeleton is therefore an under-constrained inverse problem: the measured returns constrain only part of the pose, and the remaining structure must be inferred from human-body priors.

As illustrated by the upper path in Fig.~\ref{fig:wave2body_overview}, most existing methods~\cite{m4human,wicompass,mmdiff,mmpose,mmmesh} jointly learn radar representations and human-body priors within a single supervised radar-to-coordinate regressor. Even with stronger backbones or pose-aware objectives, they typically predict raw 3D joint coordinates. Consequently, the model must use the same limited paired supervision to encode sparse radar returns, capture valid joint configurations, and align radar with body structure. This coupling is problematic because paired mmWave datasets are costly and cover limited subjects, motions, environments, and sensing conditions. Under ambiguous observations, a direct regressor may exploit training-specific sensing artifacts and pose frequencies rather than transferable radar representations and anatomical structure. Without reusable radar and body interfaces, radar representation learning, human-body structure modeling, and cross-modal alignment remain entangled within each paired training setup.

As illustrated by the lower path in Fig.~\ref{fig:wave2body_overview}, Wave2Body addresses this bottleneck by reformulating mmWave HPE as translation from a pretrained radar-token space to a pretrained body-token space. These spaces capture complementary sensor and body knowledge for representing sparse radar returns and valid human configurations. Independent pretraining creates reusable interfaces, allowing the paired training stage to focus on cross-modal mapping rather than learning both spaces from scratch.

Wave2Body realizes this formulation in stages. A self-supervised mmWave tokenizer learns from point clouds without pose supervision, while a compositional VQ-VAE body tokenizer learns from pose-only data. With both tokenizers frozen, a lightweight translator maps radar tokens to embeddings that are quantized and decoded into a pelvis-centered pose. Because body tokens exclude global position, a separate root localization head estimates the pelvis from radar observations. Only the translator and localization head use paired radar--pose supervision for cross-modal mapping and global localization, respectively.

We evaluate Wave2Body on the benchmarks of M4Human~\cite{m4human} and mmBody~\cite{mmbody} for radar-frame absolute-pose estimation using mean per-joint position error (MPJPE). Wave2Body ranks first across all reported M4Human Cross-Subject action groups, reducing micro-average MPJPE by 9.7\%; it also reduces Cross-Action and mmBody MPJPE by 3.8\% and 8.2\%. These gains support our hypothesis that decoupling radar representation learning, body modeling, and cross-modal alignment improves generalization to unseen subjects and actions. Also, Wave2Body uses \(31.00\times\) fewer training and \(89.81\times\) fewer inference FLOPs than the matched baseline.

Our contributions are summarized as follows:

\begin{itemize}
    \item We reformulate mmWave HPE as translation between independently pretrained radar- and body-token spaces, replacing direct radar-to-coordinate regression with reusable interfaces on both sides.

    \item We instantiate this formulation with a self-supervised mmWave tokenizer pretrained on radar point clouds without pose labels and a compositional body tokenizer pretrained on pose-only data; paired radar--pose supervision is used only for translation and localization.

    \item We show that this decoupled learning pipeline improves radar-frame absolute-pose generalization, especially under cross-subject and cross-action shifts, while substantially reducing training and inference FLOPs.
\end{itemize}

\section{Related Work}
\label{sec:related}

\paragraph{mmWave Human Pose Estimation.}
Early mmWave systems established the feasibility of recovering human skeletons or meshes~\cite{mmpose,mmmesh,m3track}, while datasets and benchmarks such as mRI, HuPR, mmBody, MM-Fi, RT-Pose, and M4Human enabled evaluation under more diverse sensing conditions~\cite{mri,hupr,mmbody,mmfi,rtpose,m4human}. Subsequent work explored spatiotemporal learning, radar-tensor representations, multi-view fusion, mesh reconstruction, simulation and data scaling, diffusion models, mmWave point-cloud enhancement and generation, and cross-modal learning~\cite{kong2026twr,pham2026twostage,maepose,FUSE,p4transformer,rtpose,m4esh,mmGPE,mmhpe,mmpoint,cpformer,transhupr,mvdopplerpose,millimamba,dptm,mmdiff,su2026mmwaveflow,immfusion,mmrehab,yang2026bridging}. Recent studies have highlighted the importance of pose priors and distribution-shift evaluation for generalizable mmWave HPE~\cite{mmJoints,posegraphnet,emdul,dghmesh}, coverage-aware data scaling~\cite{wicompass}, and self-supervised or synthetic-data-based wireless learning~\cite{umimo,syncheck,wiswiss}. Most existing methods, however, still learn task-specific coordinate or heatmap predictors from paired radar--pose data. Wave2Body instead changes the prediction interface, translating radar tokens into a frozen body-token space learned from broader pose data.

\paragraph{Tokenized Representations and Interface Alignment.}
Masked point-cloud pretraining methods learn local tokens by reconstructing or discriminating masked neighborhoods~\cite{pointbert,pointmae,maskpoint}, while VQ-VAE and tokenized human-pose models represent structured continuous signals through discrete, compositional latent spaces~\cite{vqvae,pct,tokenhmr,vqhps}. In parallel, multimodal models show that pretrained modality-specific interfaces can be connected through lightweight projection or cross-attention modules~\cite{flamingo,blip2,llava}. Wave2Body combines these ideas for mmWave HPE: a self-supervised mmWave tokenizer represents local radar returns, a pretrained compositional body tokenizer defines an anatomical output space, and a lightweight radar-to-body translator aligns the two spaces.

\section{Method}
\label{sec:method}

\subsection{Overview}
\label{sec:method_task}

Given a mmWave point cloud $\mathbf{X}=\{\mathbf{x}_i\}_{i=1}^{N}$, where each point $\mathbf{x}_i\in\mathbb{R}^{3+C_r}$ contains a 3D coordinate and $C_r$ radar attributes, our goal is to estimate an absolute 3D human pose $\mathbf{Y}\in\mathbb{R}^{J\times3}$ in the radar coordinate frame, with $J=22$ joints following the SMPL skeleton~\cite{smpl}.

Wave2Body first learns two reusable token interfaces independently. The mmWave tokenizer $E_{\mathrm{rad}}$ maps the point cloud to continuous radar tokens $\mathbf{S}$, whereas the body tokenizer $Q_{\mathrm{body}}\!\circ E_{\mathrm{body}}$, implemented within a VQ-VAE, maps pelvis-centered poses to discrete compositional body tokens. Paired radar--pose data are then used to learn radar-to-body token translation and global radar-frame root localization:
\begin{equation}
    \mathbf{S}=E_{\mathrm{rad}}(\mathbf{X}), \ 
    \hat{\mathbf{Z}}_e=T_\theta(\mathbf{S}), \ 
    \hat{\mathbf{B}}=D_{\mathrm{body}}\!\left(Q_{\mathrm{body}}(\hat{\mathbf{Z}}_e)\right),
    \label{eq:wave2body_token_path}
\end{equation}
where $T_\theta$ maps the radar tokens to pre-quantization body embeddings $\hat{\mathbf{Z}}_e$, and the frozen body quantizer and decoder $Q_{\mathrm{body}},D_{\mathrm{body}}$ reconstruct a pelvis-centered body configuration $\hat{\mathbf{B}}$. The localization head predicts the pelvis location $\hat{\mathbf{r}}$, yielding the radar-frame pose $\hat{\mathbf{Y}}=\hat{\mathbf{B}}+\mathbf{1}_{J}\hat{\mathbf{r}}^\top$.

\begin{figure}[t]
\centering
\captionsetup{skip=1pt}
\includegraphics[width=\columnwidth]{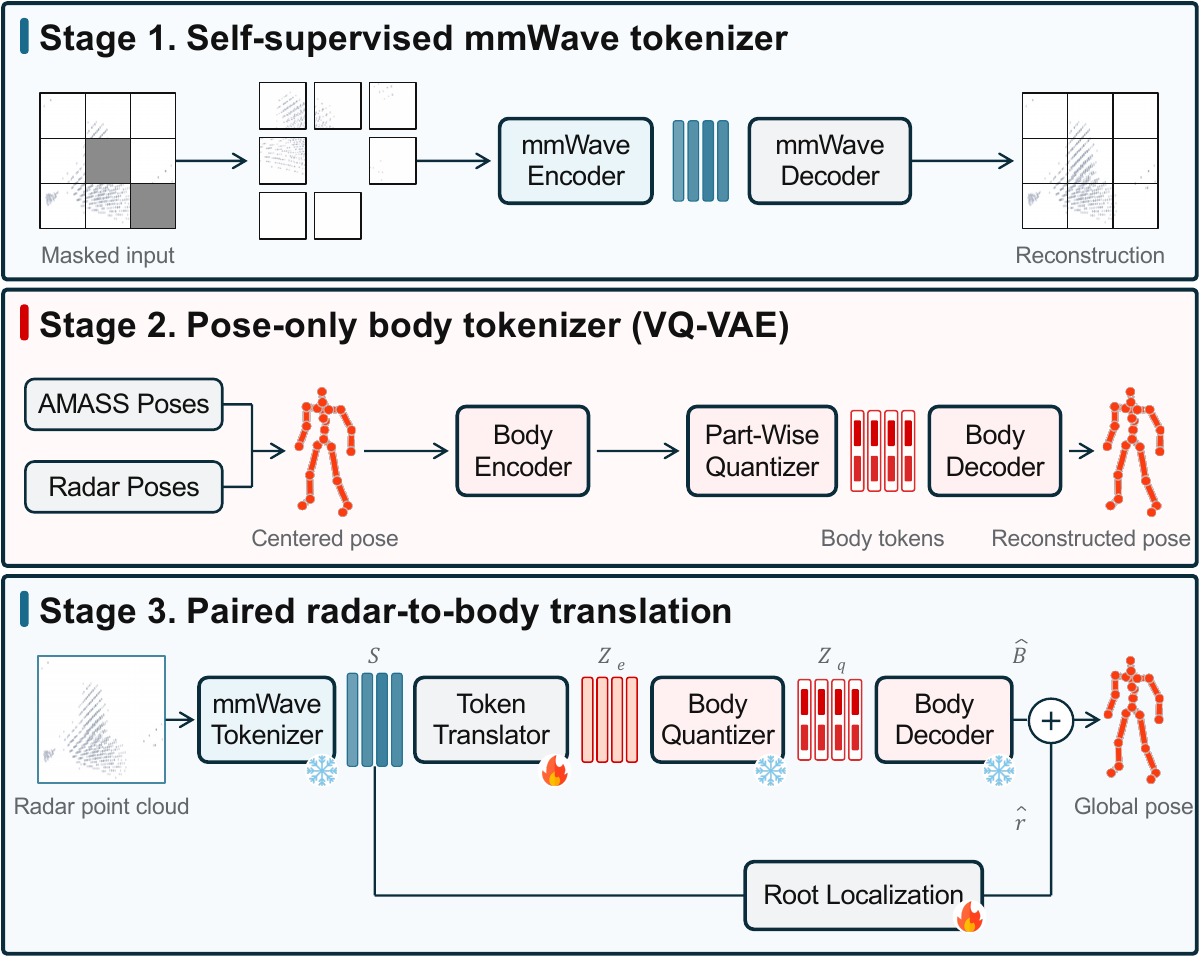}
\caption{\textbf{Three-Stage Training of Wave2Body.} Stage 1 pretrains a continuous mmWave tokenizer through masked radar-patch reconstruction. Stage 2 pretrains a discrete compositional body tokenizer, implemented as a VQ-VAE, using pose-only data from motion-capture and radar datasets. Stage 3 freezes the pretrained radar and body modules and trains the radar-to-body translator and localization head using paired radar--pose data.}
\label{fig:wave2body_training_recipe}
\vspace{-3pt}
\end{figure}

As shown in Fig.~\ref{fig:wave2body_training_recipe}, the first two stages learn the source and target token interfaces independently. Stage 1 pretrains the mmWave tokenizer without pose labels, and Stage 2 pretrains the body tokenizer from pose-only data without using radar inputs. Stage 3 freezes the pretrained radar and body modules and learns cross-modal translation and global localization from paired radar--pose data. Stage 3 is optimized in two phases: pelvis-centered token translation is learned first, followed by joint refinement of the translator and localization head under absolute-pose supervision.

\subsection{Self-Supervised mmWave Tokenizer}
\label{sec:radar_tokens}

The mmWave tokenizer $E_{\mathrm{rad}}$ is a Point-MAE-style patch encoder~\cite{pointmae} that converts a sparse, unordered point cloud into a fixed-length sequence of continuous radar tokens. We sample $G$ valid patch centers using farthest-point sampling and gather the $K$ nearest support points around each center to form local patches $\{\mathbf{P}_g\}_{g=1}^{G}$. Within each patch, spatial coordinates are expressed relative to its center $\mathbf{c}_g$, while radar attributes such as reflection intensity or velocity are retained. Each patch is embedded, augmented with a positional encoding of $\mathbf{c}_g$, and contextualized by a Transformer. The resulting sequence $\mathbf{S}=[\mathbf{s}_1,\ldots,\mathbf{s}_G]\in\mathbb{R}^{G\times d_s}$ combines centered local structure with radar-frame location.

To learn the radar-token space without pose labels, we pretrain $E_{\mathrm{rad}}$ through masked patch reconstruction. A random subset $\mathcal{M}$ of valid patches is masked, and a lightweight reconstruction decoder predicts both their local geometry and radar attributes:
\begin{equation}
    \mathcal{L}_{\mathrm{rad}}
    =
    d_{\mathrm{Chamfer}}\!\left(
        \hat{\mathbf{P}}_{\mathcal{M}}^{\mathrm{xyz}},
        \mathbf{P}_{\mathcal{M}}^{\mathrm{xyz}}
    \right)
    +
    \gamma
    \left\|
        \hat{\mathbf{P}}_{\mathcal{M}}^{\mathrm{attr}}
        -
        \mathbf{P}_{\mathcal{M}}^{\mathrm{attr}}
    \right\|_2^2 .
    \label{eq:radar_pretrain_loss}
\end{equation}
The Chamfer term measures the bidirectional distance between reconstructed and original masked point sets, while attributes are compared index-wise in KNN-distance order. After pretraining, the decoder is discarded and $E_{\mathrm{rad}}$ is frozen. All $G$ patches are then encoded to produce $\mathbf{S}$ as the source sequence for radar-to-body translation.

\subsection{Compositional Body Tokenizer}
\label{sec:body_tokens}

The body tokenizer defines a discrete target space for human articulation independently of global translation. We implement it as a part-structured VQ-VAE~\cite{vqvae} comprising a body encoder $E_{\mathrm{body}}$, quantizer $Q_{\mathrm{body}}$, and decoder $D_{\mathrm{body}}$. Whereas the mmWave tokenizer produces continuous radar tokens, the body tokenizer $Q_{\mathrm{body}}\!\circ E_{\mathrm{body}}$ assigns pose embeddings to compositional codebook entries. Because paired mmWave datasets typically contain limited pose diversity, we train the body tokenizer using large-scale motion-capture poses together with pose annotations from the corresponding radar benchmark train split, without using their paired radar inputs. All poses are converted to the same 22-joint topology and centered at the pelvis:
\begin{equation}
    \mathbf{B}=\mathbf{Y}-\mathbf{1}_{J}\mathbf{r}^{\top},
    \qquad \mathbf{B}\in\mathbb{R}^{J\times3},
    \label{eq:body_configuration}
\end{equation}
where $\mathbf{r}\in\mathbb{R}^3$ is the pelvis coordinate.

We train the body-token VQ-VAE to define a structured prediction space rather than to serve as a motion generator. To make the tokens compositional, we divide the skeleton into five kinematic regions: left leg (left hip/knee/ankle/foot), right leg (right hip/knee/ankle/foot), spine core (spine1--3, neck, and head), left arm (spine3 and left collar/shoulder/elbow/wrist), and right arm (spine3 and right collar/shoulder/elbow/wrist). Each region has a part-specific encoder, codebook, and decoder, with spine3 shared by the torso and arm regions. Collectively, the encoders produce $T=24$ pre-quantization embeddings $\mathbf{Z}_e=E_{\mathrm{body}}(\mathbf{B})$, allocated as 5, 5, 4, 5, and 5 slots across the five regions. For embedding $\mathbf{z}_{p,t}$ at slot $t$ of region $p$, quantization selects the nearest entry in its codebook $\mathcal{C}_p$:
\begin{equation}
    Q_p(\mathbf{z}_{p,t})=\mathbf{e}^{*}, \ 
    \mathbf{e}^{*}
    =
    \arg\min_{\mathbf{e}\in\mathcal{C}_p}
    \left\|\mathbf{z}_{p,t}-\mathbf{e}\right\|_2^2 .
    \label{eq:part_quantization}
\end{equation}
The selected codebook entry $\mathbf{e}^{*}$ is a quantized body token, and its codebook index is the corresponding discrete token ID. The quantized tokens from all regions are concatenated as $\mathbf{Z}_q=Q_{\mathrm{body}}(\mathbf{Z}_e)$. Part-specific decoders reconstruct their corresponding joint subsets; predictions for shared joints are averaged, and the pelvis remains fixed at the origin. The VQ-VAE is optimized with coordinate reconstruction and vector-quantization objectives:
\begin{equation}
    \mathcal{L}_{\mathrm{body}}
    =
    \ell_{\mathrm{rec}}\!\left(
        D_{\mathrm{body}}\!\left(
            Q_{\mathrm{body}}\!\left(E_{\mathrm{body}}(\mathbf{B})\right)
        \right),
        \mathbf{B}
    \right)
    +
    \ell_{\mathrm{vq}} .
    \label{eq:body_vq_loss}
\end{equation}

After pretraining, $E_{\mathrm{body}}$, $Q_{\mathrm{body}}$, and $D_{\mathrm{body}}$ are frozen. During translator training, the body encoder supplies target pre-quantization embeddings, while the quantizer and decoder map the translator outputs through the discrete body-token space to a reconstructed pose. The radar model predicts through a body representation learned from broad pose data rather than regressing joint coordinates.

\subsection{Radar-to-Body Token Translation and Localization}
\label{sec:radar_to_body_translation}

Given the radar-token sequence $\mathbf{S}$, the query-based Transformer $T_\theta$ predicts pre-quantization body embeddings $\hat{\mathbf{Z}}_e\in\mathbb{R}^{T\times d_b}$. It contains $T$ learnable body queries, one for each body-token slot. Each query cross-attends to the radar tokens, after which self-attention models dependencies among body parts. The predicted embeddings are quantized as $\hat{\mathbf{Z}}_q=Q_{\mathrm{body}}(\hat{\mathbf{Z}}_e)$ and decoded according to Eq.~\ref{eq:wave2body_token_path}. The body queries therefore act as anatomical output slots, allowing weakly observed limbs to be inferred jointly from the radar-token sequence and the learned body structure.

Let $\mathbf{Z}_e=E_{\mathrm{body}}(\mathbf{B})$ denote the target pre-quantization body embeddings. The translator is trained with an embedding-alignment loss and a reconstruction loss on the decoded pelvis-centered pose:
\begin{equation}
    \mathcal{L}_{\mathrm{trans}}
    =
    \lambda_z
    \left\|
        \hat{\mathbf{Z}}_e-\operatorname{sg}(\mathbf{Z}_e)
    \right\|_2^2
    +
    \lambda_b
    \left\|
        \hat{\mathbf{B}}-\mathbf{B}
    \right\|_F^2 ,
    \label{eq:translator_loss}
\end{equation}
where $\operatorname{sg}(\cdot)$ denotes stop-gradient. Gradients through the quantizer are estimated with the straight-through estimator.

Because the body-token space excludes global translation, the lightweight query-attention localization head separately predicts the pelvis location. Its inputs are the radar tokens $\{\mathbf{s}_g\}_{g=1}^{G}$, their patch centers $\{\mathbf{c}_g\}_{g=1}^{G}$, and compact valid-point statistics $\rho(\mathbf{X})$. In our implementation, $\rho(\mathbf{X})$ contains global xyz mean, standard deviation, bounding-box extrema, intensity mean, intensity standard deviation, intensity extrema, and the intensity-magnitude-weighted xyz mean. We use the latter as the root prior $\mathbf{r}_0$. Each radar token is augmented with its patch center and its offset from this prior:
\begin{equation}
    \mathbf{u}_g=
    \left[
        \mathbf{s}_g,\,
        \mathbf{c}_g,\,
        \mathbf{c}_g-\mathbf{r}_0
    \right].
    \label{eq:root_token}
\end{equation}
Let $\mathbf{U}=\{\mathbf{u}_g\}_{g=1}^{G}$ and $\mathbf{C}=\{\mathbf{c}_g\}_{g=1}^{G}$. Learnable root queries $\mathbf{Q}_r$ attend to $\mathbf{U}$, and their pooled output is combined with $\rho(\mathbf{X})$ and summaries $\sigma(\mathbf{S},\mathbf{C})$, comprising mean- and max-pooled radar tokens and the mean, minimum, and maximum patch centers, to predict a residual:
\begin{equation}
    \begin{aligned}
        \mathbf{h}_r
        &=\operatorname{Pool}\!\left(\operatorname{Attn}(\mathbf{Q}_r,\mathbf{U})\right),\\
        \Delta\hat{\mathbf{r}}
        &=R_\phi\!\left(
            [\mathbf{h}_r,\sigma(\mathbf{S},\mathbf{C}),\rho(\mathbf{X})]
        \right),\\
        \hat{\mathbf{r}}
        &=\mathbf{r}_0+\Delta\hat{\mathbf{r}}.
    \end{aligned}
    \label{eq:root_head}
\end{equation}

This design keeps radar-frame localization tied to absolute sensor geometry, while the body-token path predicts pelvis-centered articulation. During the final optimization phase, $E_{\mathrm{rad}}$, $E_{\mathrm{body}}$, $Q_{\mathrm{body}}$, and $D_{\mathrm{body}}$ remain frozen, while $T_\theta$ and $R_\phi$ are jointly optimized with
\begin{equation}
    \mathcal{L}_{\mathrm{final}}
    =
    \mathcal{L}_{\mathrm{trans}}
    +\lambda_r\left\|\hat{\mathbf{r}}-\mathbf{r}\right\|_2^2
    +\lambda_y\left\|\hat{\mathbf{Y}}-\mathbf{Y}\right\|_F^2 .
    \label{eq:final_loss}
\end{equation}
Here, $\mathcal{L}_{\mathrm{trans}}$ provides pre-quantization embedding and pelvis-centered pose supervision, while the remaining terms supervise root localization and radar-frame absolute pose.

\section{Experiments}
\label{sec:experiments}

\begin{table*}[t]
\centering
\caption{\textbf{Radar-frame absolute-pose estimation across benchmarks.} Absolute MPJPE (mm; lower is better) on M4Human and mmBody. For M4Human, Micro Average is sample-weighted, and Cross-Action groups contain only held-out test actions. Each learned entry is reported as the mean \(\pm\) standard deviation over three seeds. Bold and underline indicate the best and second-best results.}
\label{tab:absolute_baselines}
\setlength{\tabcolsep}{3.6pt}
\footnotesize
\begin{tabular}{llrrrrr}
\toprule
Dataset / Split & Action Group & \textbf{Wave2Body (ours)} \(\downarrow\) & WiCompass~\cite{wicompass}\textsuperscript{*} \(\downarrow\) & RT-Mesh~\cite{m4human} \(\downarrow\) & mmMesh~\cite{mmmesh}\textsuperscript{*} \(\downarrow\) & mmDiff~\cite{mmdiff}\textsuperscript{*} \(\downarrow\) \\
\midrule
M4Human Cross-Subject & In-Place & \textbf{90.62 {\scriptsize\(\pm0.74\)} {\scriptsize(-10.1\%)}} & \underline{100.83 {\scriptsize\(\pm0.73\)}} & 109.60 & 128.96 {\scriptsize\(\pm2.71\)} & 131.58 {\scriptsize\(\pm5.33\)} \\
 & Sit-In-Place & \textbf{99.30 {\scriptsize\(\pm2.54\)} {\scriptsize(-15.4\%)}} & \underline{117.44 {\scriptsize\(\pm2.53\)}} & 130.80 & 162.65 {\scriptsize\(\pm2.67\)} & 158.11 {\scriptsize\(\pm13.50\)} \\
 & Non-In-Place & \textbf{127.62 {\scriptsize\(\pm0.64\)} {\scriptsize(-7.5\%)}} & \underline{137.91 {\scriptsize\(\pm4.20\)}} & 151.80 & 175.33 {\scriptsize\(\pm2.57\)} & 182.60 {\scriptsize\(\pm8.13\)} \\
 & Micro Average & \textbf{102.53 {\scriptsize\(\pm0.29\)} {\scriptsize(-9.7\%)}} & \underline{113.53 {\scriptsize\(\pm1.91\)}} & 120.20 & 146.10 {\scriptsize\(\pm2.46\)} & 149.42 {\scriptsize\(\pm5.32\)} \\
\midrule
M4Human Cross-Action & In-Place & \underline{97.30 {\scriptsize\(\pm0.66\)} {\scriptsize(+3.6\%)}} & \textbf{93.92 {\scriptsize\(\pm2.18\)}} & 107.20 & 129.32 {\scriptsize\(\pm1.09\)} & 138.71 {\scriptsize\(\pm3.70\)} \\
 & Sit-In-Place & \textbf{118.29 {\scriptsize\(\pm4.30\)} {\scriptsize(-1.4\%)}} & 125.77 {\scriptsize\(\pm1.69\)} & \underline{120.00} & 122.51 {\scriptsize\(\pm1.68\)} & 151.82 {\scriptsize\(\pm15.10\)} \\
 & Non-In-Place & \textbf{137.26 {\scriptsize\(\pm0.75\)} {\scriptsize(-9.2\%)}} & \underline{151.16 {\scriptsize\(\pm8.71\)}} & 152.60 & 183.84 {\scriptsize\(\pm1.64\)} & 193.12 {\scriptsize\(\pm3.19\)} \\
 & Micro Average & \textbf{115.26 {\scriptsize\(\pm0.98\)} {\scriptsize(-3.8\%)}} & \underline{119.81 {\scriptsize\(\pm4.36\)}} & 122.00 & 150.34 {\scriptsize\(\pm1.03\)} & 161.64 {\scriptsize\(\pm4.55\)} \\
\midrule
mmBody & All & \textbf{128.17 {\scriptsize\(\pm1.87\)} {\scriptsize(-8.2\%)}} & \underline{139.69 {\scriptsize\(\pm4.43\)}} & -- & 145.79 {\scriptsize\(\pm2.91\)} & 267.37 {\scriptsize\(\pm19.36\)} \\
\bottomrule
\end{tabular}
\begin{flushleft}
\scriptsize
\textsuperscript{*}Re-evaluated under our radar-frame absolute-pose protocol; original metrics may use different alignments. RT-Mesh uses official M4Human results. Parentheses show relative change from the best competitor; negative values indicate improvement.
\end{flushleft}
\end{table*}

\begin{table*}[t]
\centering
\caption{\textbf{M4Human absolute-pose diagnostics.} Errors (mm; lower is better) for Wave2Body and the matched WiCompass baseline under Cross-Subject and Cross-Action shifts. Each entry is reported as the mean \(\pm\) standard deviation over three seeds; bold indicates the better result.}
\label{tab:m4human_absolute_main}
\setlength{\tabcolsep}{1.8pt}
\scriptsize
\begin{tabular*}{\textwidth}{@{\extracolsep{\fill}}llcccccccc@{}}
\toprule
& & \multicolumn{2}{c}{Absolute MPJPE \(\downarrow\)} & \multicolumn{2}{c}{Root error \(\downarrow\)} & \multicolumn{2}{c}{Root-aligned MPJPE \(\downarrow\)} & \multicolumn{2}{c}{PA-MPJPE \(\downarrow\)} \\
\cmidrule(lr){3-4}\cmidrule(lr){5-6}\cmidrule(lr){7-8}\cmidrule(lr){9-10}
Split & Action Group & \wbt{Wave2Body} & \wct{WiCompass} & \wbt{Wave2Body} & \wct{WiCompass} & \wbt{Wave2Body} & \wct{WiCompass} & \wbt{Wave2Body} & \wct{WiCompass} \\
\midrule
Cross-Subject & In-Place & \wbt{\textbf{90.62 {\tiny\(\pm0.74\)}}} & \wct{100.83 {\tiny\(\pm0.73\)}} & \wbt{\textbf{67.76 {\tiny\(\pm0.80\)}}} & \wct{78.31 {\tiny\(\pm0.89\)}} & \wbt{\textbf{66.80 {\tiny\(\pm0.20\)}}} & \wct{72.65 {\tiny\(\pm1.59\)}} & \wbt{\textbf{50.83 {\tiny\(\pm0.07\)}}} & \wct{53.42 {\tiny\(\pm0.35\)}} \\
 & Sit-In-Place & \wbt{\textbf{99.30 {\tiny\(\pm2.54\)}}} & \wct{117.44 {\tiny\(\pm2.53\)}} & \wbt{\textbf{85.16 {\tiny\(\pm3.41\)}}} & \wct{98.94 {\tiny\(\pm4.30\)}} & \wbt{\textbf{64.14 {\tiny\(\pm0.20\)}}} & \wct{77.57 {\tiny\(\pm0.89\)}} & \wbt{\textbf{48.69 {\tiny\(\pm0.17\)}}} & \wct{56.51 {\tiny\(\pm0.89\)}} \\
 & Non-In-Place & \wbt{\textbf{127.62 {\tiny\(\pm0.64\)}}} & \wct{137.91 {\tiny\(\pm4.20\)}} & \wbt{\textbf{88.20 {\tiny\(\pm0.60\)}}} & \wct{105.57 {\tiny\(\pm4.46\)}} & \wbt{\textbf{95.40 {\tiny\(\pm0.31\)}}} & \wct{95.54 {\tiny\(\pm1.54\)}} & \wbt{\textbf{68.36 {\tiny\(\pm0.03\)}}} & \wct{68.44 {\tiny\(\pm0.49\)}} \\
\midrule
Cross-Action & In-Place & \wbt{97.30 {\tiny\(\pm0.66\)}} & \wct{\textbf{93.92 {\tiny\(\pm2.18\)}}} & \wbt{62.23 {\tiny\(\pm0.48\)}} & \wct{\textbf{60.31 {\tiny\(\pm0.77\)}}} & \wbt{75.64 {\tiny\(\pm0.46\)}} & \wct{\textbf{74.59 {\tiny\(\pm2.02\)}}} & \wbt{67.64 {\tiny\(\pm0.46\)}} & \wct{\textbf{63.83 {\tiny\(\pm1.42\)}}} \\
 & Sit-In-Place & \wbt{\textbf{118.29 {\tiny\(\pm4.30\)}}} & \wct{125.77 {\tiny\(\pm1.69\)}} & \wbt{\textbf{84.41 {\tiny\(\pm6.58\)}}} & \wct{102.62 {\tiny\(\pm11.73\)}} & \wbt{89.93 {\tiny\(\pm2.75\)}} & \wct{\textbf{81.14 {\tiny\(\pm1.89\)}}} & \wbt{73.09 {\tiny\(\pm0.72\)}} & \wct{\textbf{67.33 {\tiny\(\pm2.36\)}}} \\
 & Non-In-Place & \wbt{\textbf{137.26 {\tiny\(\pm0.75\)}}} & \wct{151.16 {\tiny\(\pm8.71\)}} & \wbt{\textbf{91.10 {\tiny\(\pm0.79\)}}} & \wct{108.88 {\tiny\(\pm7.69\)}} & \wbt{\textbf{100.84 {\tiny\(\pm0.41\)}}} & \wct{102.06 {\tiny\(\pm4.17\)}} & \wbt{\textbf{78.68 {\tiny\(\pm0.09\)}}} & \wct{78.93 {\tiny\(\pm2.12\)}} \\
\bottomrule
\end{tabular*}
\end{table*}

\subsection{Experimental Setup}

\paragraph{Benchmarks and Protocols.}
We evaluate Wave2Body on M4Human~\cite{m4human} and mmBody~\cite{mmbody}. M4Human is used for generalization evaluation under its official Cross-Subject and Cross-Action protocols, which hold out identities and action classes, respectively. We omit the Random split because it shares both subjects and action classes between training and test data and mainly reflects in-distribution interpolation. Following the benchmark protocol, we report In-Place, Sit-In-Place, Non-In-Place, and a sample-weighted micro-average over all actions. On mmBody, we use the official train/test partition to assess whether the gains persist on a second benchmark, training a model for each benchmark and split.

\paragraph{Input Preprocessing.}
For Wave2Body, M4Human inputs contain $N=1024$ points with xyz coordinates and reflection intensity, causally aggregated from the current and previous three frames; mmBody inputs contain $N=200$ points with xyz, intensity, and velocity over three frames. The reimplemented baselines operate on the same official examples, splits, radar point-cloud modality, coordinate frame, and pose targets.

\paragraph{Pose Representation and Evaluation.}
All annotations are mapped to a common SMPL-style 22-joint skeleton and evaluated in the radar coordinate frame. The primary metric is absolute MPJPE in millimeters, which measures both body articulation and global pelvis localization. We additionally report root error, root-aligned MPJPE, and Procrustes-aligned MPJPE (PA-MPJPE). Unless noted otherwise, learned entries report the mean and sample standard deviation over seeds 42, 43, and 44; for each seed, the validation-selected checkpoint is evaluated on the test split.

\paragraph{Wave2Body Implementation.}
Wave2Body follows the three-stage training pipeline described in Sec.~\ref{sec:method}. The mmWave tokenizer uses $G=96$ patch centers selected by farthest-point sampling and $K=32$ nearest neighbors per patch, and outputs 128-D radar tokens; it is pretrained for 80 epochs with 60\% patch masking. The part-level body tokenizer is trained for 30 epochs on pelvis-centered AMASS~\cite{mahmood2019amass} poses and the corresponding benchmark training poses, using 24 latent slots, five kinematic regions (left leg, right leg, spine core, left arm, and right arm), 256-D embeddings, and 96-entry codebooks. In Stage 3, we first train the radar-to-body translator for 25 epochs with all tokenizer modules frozen, then add the localization head and jointly train it with the translator for another 25 epochs. The translator uses 24 body queries, four cross-attention blocks, and two self-attention blocks. All stages use AdamW and bf16 mixed precision. No component is trained on test samples; checkpoints are selected using the official M4Human validation split or held-out training data.

\paragraph{Baselines.}
We compare with WiCompass~\cite{wicompass}, mmMesh~\cite{mmmesh}, and mmDiff~\cite{mmdiff} under a matched point-cloud protocol. These baselines use the same paired radar--pose splits, validation and test protocol, radar-frame 22-joint targets, and evaluation code as Wave2Body, but retain architecture-specific input preprocessing. They directly predict absolute joint coordinates. We also report RT-Mesh~\cite{m4human}, the official M4Human 4D radar-tensor baseline, but exclude it on mmBody because mmBody provides only point clouds. We do not mix root-relative results from prior reports with our absolute-coordinate evaluation. The only additional supervision used by Wave2Body is AMASS pose-only data for body-tokenizer pretraining.

\begin{figure}[t]
\centering
\captionsetup{skip=1pt}
\includegraphics[width=\columnwidth]{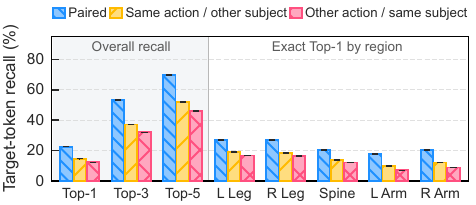}
\caption{\textbf{Body-token accuracy.} M4Human Cross-Subject token-ID accuracy. Error bars show standard deviations across three seeds.}
\label{fig:token_alignment_m4human}
\vspace{-12pt}
\end{figure}

\subsection{Main Results} \label{sec:main_results}

\paragraph{Pose Accuracy.}
Table~\ref{tab:absolute_baselines} reports absolute MPJPE by split and action group. Wave2Body achieves the best overall result in seven of eight M4Human cross-domain rows: all Cross-Subject groups and the Cross-Action Sit-In-Place, Non-In-Place, and micro-average rows. It reduces the Cross-Subject and Cross-Action micro-averages by 9.7\% and 3.8\%, respectively, and mmBody MPJPE by 8.2\%. Cross-Action results are mixed: it lowers Non-In-Place error by 9.2\% but is 3.6\% worse than WiCompass on In-Place actions; its 3.8\% micro-average gain also trails its 9.7\% Cross-Subject gain. Benefits thus concentrate on subject shift and dynamic held-out motions, suggesting a trade-off between a discrete body prior's robustness and direct coordinate regression's fine-grained precision (Sec.~\ref{sec:discussion}).

\begin{figure}[t]
\centering
\captionsetup{skip=1pt}
\includegraphics[width=\columnwidth]{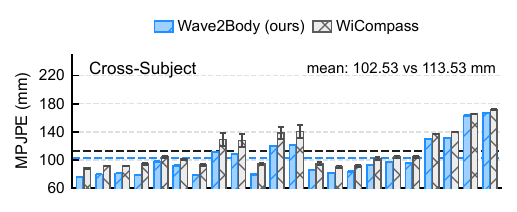}
\includegraphics[width=\columnwidth]{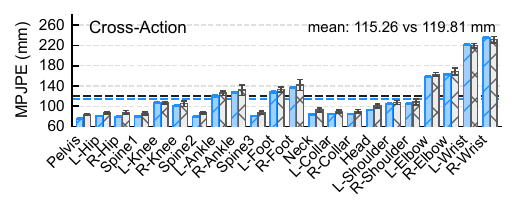}
\caption{\textbf{M4Human per-joint error.}  Error bars show three-seed standard deviations; dashes mark the mean.}
\label{fig:per_joint_m4human}
\vspace{-4pt}
\end{figure}
\vspace{-8pt}

\paragraph{Error Decomposition.}
Table~\ref{tab:m4human_absolute_main} decomposes absolute-pose error into global localization and body-configuration components. Under Cross-Subject, Wave2Body improves all four metrics in every action group, so its absolute-pose gains extend across both localization and articulation. Under Cross-Action, the pattern is more localized: Wave2Body reduces absolute and root errors for Sit-In-Place and Non-In-Place, while the root-aligned and Procrustes-aligned results are mixed. Only Non-In-Place improves across all four metrics. Thus, the cross-action gains arise primarily from stronger root localization, whereas the articulation benefit is clearest under subject shift and dynamic held-out actions.

\begin{figure*}[t]
\centering
\captionsetup{skip=1pt}
\includegraphics[width=\textwidth]{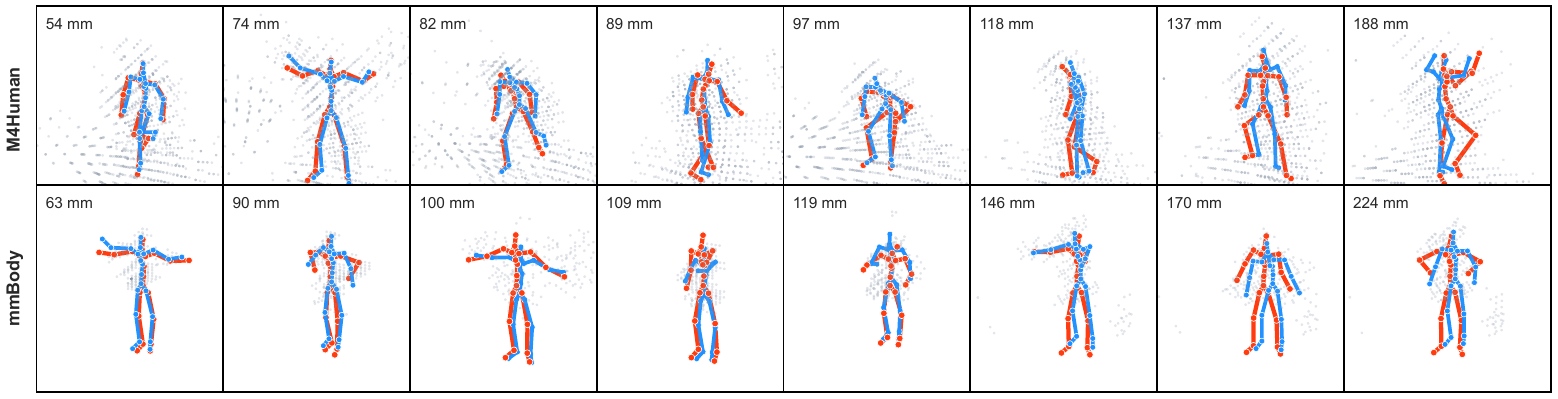}
\caption{\textbf{Qualitative results from easy to challenging cases.} Representative M4Human and mmBody test samples are selected across the absolute-MPJPE spectrum. Gray points denote the input mmWave point cloud, red skeletons denote ground truth, and blue skeletons denote Wave2Body predictions. The value in each panel is the sample-level radar-frame absolute MPJPE.}
\label{fig:qualitative_error_buckets}
\vspace{-3pt}
\end{figure*}

\begin{figure*}[t]
\centering
\captionsetup{skip=1pt}
\includegraphics[width=\textwidth]{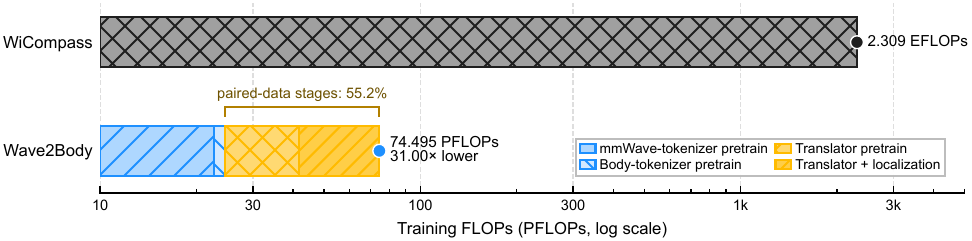}
\caption{\textbf{Cold-start training computation on M4Human.} Wave2Body is decomposed into two tokenizer-pretraining phases and two paired-data phases. FLOPs include supported profiler operations in forward, loss, backward, and optimizer computation.}
\label{fig:efficiency_training}
\vspace{-4pt}
\end{figure*}

\paragraph{Modality Alignment Analysis.}
We test whether the translator maps mmWave representations into the body-token space~\cite{duan2022codebook}. Low end-to-end MPJPE along the radar-to-token-to-pose path shows accurate recovery but cannot rule out frequent-pose shortcuts. We therefore evaluate the translator-only checkpoint before localization. On each M4Human test split, Top-\(k\) recall~\cite{lee2018scan} asks whether the paired ground-truth token at each of 24 slots is among the \(k\) pose codes nearest the predicted embedding; Top-1 is exact token agreement. To test sample specificity, we use two deterministic one-to-one controls: same action/other subject and other action/same subject. Across three seeds on Cross-Subject, paired Top-1/3/5 recall is 22.63/53.16/69.77\%, exceeding the 14.68/37.20/52.12\% and 12.25/32.02/45.98\% controls (Fig.~\ref{fig:token_alignment_m4human}); the advantage also holds in all five regions, supporting sample-specific recovery. On Cross-Action, the paired and control Top-1 recalls are 21.60\%, 12.49\%, and 10.80\%, respectively. We then test whether token alignment tracks articulation error. For each sample, we correlate the Top-\(k\) miss rate (fraction of 24 targets absent from Top-\(k\)) with root-aligned MPJPE using Spearman's rank correlation \(\rho\)~\cite{spearman1904}. On Cross-Subject, Top-1/3/5 mean correlations are \(\rho=0.595\), \(\rho=0.742\), and \(\rho=0.774\); Cross-Action Top-1 is \(\rho=0.615\). Under Schober et al.'s scale~\cite{schober2018}, the Top-1 correlation is moderate, while the Top-3/5 correlations are strong; token-alignment mismatch correlates with predicted-pose error. Together, low end-to-end error, stronger paired-than-control recovery, and this sample-level error link support alignment beyond pose-frequency, action, or identity shortcuts.

\paragraph{Qualitative Results.}
Figure~\ref{fig:qualitative_error_buckets} complements aggregate results with representative M4Human and mmBody predictions selected from low- to high-error buckets. At low and moderate error levels, predictions closely follow ground-truth torso and limb configurations despite sparse, partial radar coverage. In the hardest cases, Wave2Body still tends to produce human-like skeletons, but errors manifest as global root shifts or ambiguous distal-limb articulation. Such failures are not caused only by sparse points: dynamic motion, unusual poses, and difficult scenarios can make short-window radar observations compatible with multiple plausible poses. This supports the decomposition in Table~\ref{tab:m4human_absolute_main}: decoupling improves average articulation and localization under shift, but sparse, cluttered, or ambiguous returns can still yield plausible but incorrect body tokens.

\begin{table}[t]
\centering
\caption{\textbf{Inference efficiency on M4Human.} Measurements are CUDA-synchronized and exclude data loading and transfer.}
\label{tab:efficiency_inference}
\setlength{\tabcolsep}{3.5pt}
\resizebox{\columnwidth}{!}{%
\begin{tabular}{lrrrr}
\toprule
Method & Params & GFLOPs/sample & Throughput & Peak CUDA Mem. \\
& & & (samples/s, B=192) & (MiB, B=192) \\
\midrule
WiCompass & 26.91M & 62.381 & 536.6 & 7556.4 \\
Wave2Body & 16.93M & 0.6946 & 4328.5 & 558.9 \\
\bottomrule
\end{tabular}
}
\end{table}

\vspace{-8pt}

\begin{figure*}[t]
\centering
\captionsetup{skip=1pt}
\includegraphics[width=\textwidth]{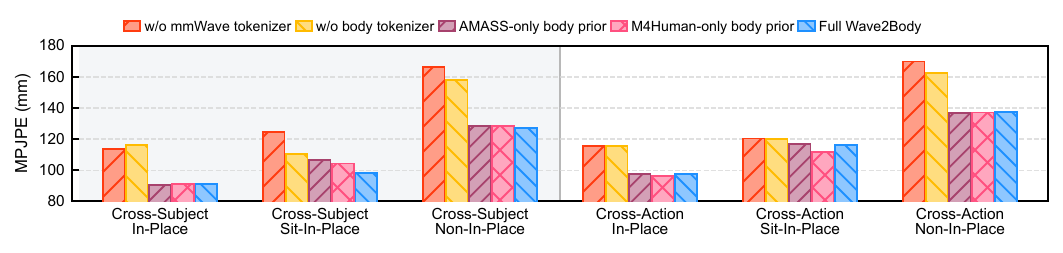}
\caption{\textbf{Component ablations studies.} Lower MPJPE is better.}
\label{fig:ablation_test_abs_mpjpe}
\vspace{-4pt}
\end{figure*}
\vspace{-8pt}

\paragraph{Per-Joint Error Analysis.}\label{sec:per_joint_error}
We analyze absolute-pose error across body joints on M4Human against WiCompass, the strongest matched baseline (Table~\ref{tab:absolute_baselines}). Dashed lines in Fig.~\ref{fig:per_joint_m4human} show the unweighted joint mean. Under Cross-Subject, Wave2Body improves every joint and lowers mean error by 9.7\%, with the largest gains at ankles, feet, spine, pelvis, and hips, indicating better global placement and lower-body localization under subject shift. Under Cross-Action, it lowers the mean by 3.8\% and improves 19 of 22 joints, most clearly at the pelvis, hips, spine, neck, and collar. In both splits, wrists, elbows, ankles, and feet have the highest errors, consistent with distal joints being less constrained by mmWave reflections than the pelvis and torso.

\subsection{Efficiency Analysis} \label{sec:efficiency}

\paragraph{Profiling Protocol.}
We profile Wave2Body and WiCompass, the strongest matched baseline, on the M4Human Cross-Subject full-action setting using a single NVIDIA GeForce RTX 4090 with bf16 mixed precision, TF32, PyTorch eager execution, and \texttt{torch.compile} disabled. For each training phase, we average supported-operation FLOPs over 128 batches, covering forward propagation, loss computation, backward propagation, and optimizer updates, and scale the per-sample average by the training samples per epoch and configured epochs. Training batch sizes are 192 for Wave2Body and 48 for WiCompass due to GPU memory limits. Data loading, validation, and checkpointing are excluded.

\paragraph{Training Computation.}
Figure~\ref{fig:efficiency_training} compares total training computation and decomposes Wave2Body by phase. Under the same accounting convention, Wave2Body uses \(31.00\times\) fewer cold-start training FLOPs than WiCompass. Its two paired-data stages account for 55.2\% of the cold-start computation, while the remaining 44.8\% is spent on pretraining the two tokenizers.

\paragraph{Inference Efficiency.}
Table~\ref{tab:efficiency_inference} reports inference-time parameters and supported-operation FLOPs together with CUDA-synchronized throughput and peak allocated GPU memory. Compared with WiCompass, Wave2Body uses 37\% fewer parameters and \(89.81\times\) fewer forward FLOPs; at batch size 192, it achieves \(8.07\times\) higher throughput and reduces peak memory by 92.6\%. These results demonstrate substantially lower inference computation and resource use for high-throughput deployment scenarios.

\subsection{Ablation Studies} \label{sec:ablations}

We ablate three design factors behind Wave2Body: the mmWave tokenizer, the body tokenizer, and the pose data used to train the body tokenizer. Figure~\ref{fig:ablation_test_abs_mpjpe} compares the full model with four single-seed variants on M4Human Cross-Subject and Cross-Action, which evaluate held-out identities and held-out action classes, respectively. All tokenizer pretraining remains restricted to the training split.

\paragraph{Dual-Tokenizer Ablation.}
The variant without the body tokenizer retains the frozen mmWave tokenizer and localization head, but replaces body-token translation and decoding with a pooled regressor for pelvis-centered joint coordinates. It increases MPJPE in every action group by 3.3--27.5\%. Conversely, the variant without the mmWave tokenizer replaces masked-pretrained radar patches with a supervised raw-point encoder while retaining the body-token target and localization head; it increases MPJPE by 3.7--30.7\%, with the largest penalty on Cross-Subject Non-In-Place. The comparable degradation from either removal supports the dual-interface formulation: the mmWave tokenizer contributes transferable sensor representations, while the body tokenizer supplies an anatomical target space.

\paragraph{Body-Prior Training Data.}
We further train the body tokenizer using only M4Human poses, only AMASS poses, or their combination. All three variants remain stronger than removing either tokenizer. The mixed AMASS+M4Human prior performs best on the Cross-Subject micro-average and several Cross-Subject groups, whereas the M4Human-only and AMASS-only variants are each slightly better on individual Cross-Action groups.

\section{Discussion}
\label{sec:discussion}

Wave2Body does not eliminate the sensing gap between mmWave radar and vision-, depth-, or LiDAR-based sensors: radar point clouds remain sparse, specular, and viewpoint-dependent, so many body parts may be weakly observed or missing. Instead, it handles this ambiguity across three components: the mmWave tokenizer learns sensor-side regularities from unlabeled radar, the body-token space provides a structured target for articulation from pose-only data, and the localization head recovers global position. This decomposition allows each data source to contribute without requiring all of this knowledge to be learned from paired radar--pose samples~\cite{emdul,wicompass}. The results, however, show that its benefits are not uniform. Wave2Body improves consistently under Cross-Subject shift, whereas its Cross-Action gains are smaller, arise primarily from root localization, and do not extend to In-Place actions. This suggests a potential trade-off from the discrete body-token space: it constrains ambiguous predictions to plausible body configurations, but quantization may limit fine-grained accuracy when radar observations already provide sufficient pose cues. This may explain why Wave2Body is strongest under subject shift and dynamic held-out motions but weaker on Cross-Action In-Place.

\paragraph{Limitations.}
Wave2Body still depends on the quality of radar observations. When point clouds are extremely sparse, biased, or cluttered, the translated embeddings may be quantized into plausible but incorrect body tokens, and the localization head may fail when global localization cues are weak. Future work can improve robustness through explicit sequence modeling beyond fixed short-window aggregation, together with motion smoothness and kinematic consistency constraints, so that weakly observed body parts and root trajectories can be inferred over longer sequences. Another important direction is multi-person mmWave HPE~\cite{zhang2026mupose}, where inter-person occlusion and identity association make both body-token selection and radar-frame localization more challenging.

\section{Conclusion}
\label{sec:conclusion}

Wave2Body reformulates mmWave human pose estimation as radar-to-body token translation, connecting continuous tokens from a self-supervised mmWave tokenizer to the discrete space of a pretrained compositional body tokenizer. This decoupled learning pipeline gives each data source a clear role: unlabeled radar data for sensor representation learning, pose-only motion data for human-body priors, and paired radar--pose samples for cross-modal alignment. This allocation improves radar-frame pose generalization and suggests that structured prediction spaces can bridge radar observations and transferable human knowledge.

{
    \small
    \bibliographystyle{ieeenat_fullname}
    \bibliography{references}

@inproceedings{mmGPE,
  title={Towards generalized mmwave-based human pose estimation through signal augmentation},
  author={Xue, Hongfei and Cao, Qiming and Miao, Chenglin and Ju, Yan and Hu, Haochen and Zhang, Aidong and Su, Lu},
  booktitle={Proceedings of the 29th Annual International Conference on Mobile Computing and Networking},
  pages={1--15},
  year={2023}
}

@inproceedings{FUSE,
  title={Fast and scalable human pose estimation using mmwave point cloud},
  author={An, Sizhe and Ogras, Umit Y},
  booktitle={Proceedings of the 59th ACM/IEEE Design Automation Conference},
  pages={889--894},
  year={2022}
}

@article{spearman1904,
  title={The proof and measurement of association between two things.},
  author={Spearman, Charles},
  year={1961},
  publisher={Appleton-Century-Crofts}
}

@article{schober2018,
  title={Correlation coefficients: appropriate use and interpretation},
  author={Schober, Patrick and Boer, Christa and Schwarte, Lothar A},
  journal={Anesthesia \& analgesia},
  volume={126},
  number={5},
  pages={1763--1768},
  year={2018},
  publisher={LWW}
}

@inproceedings{mmdiff,
  title={Diffusion model is a good pose estimator from 3d rf-vision},
  author={Fan, Junqiao and Yang, Jianfei and Xu, Yuecong and Xie, Lihua},
  booktitle={European Conference on Computer Vision},
  pages={1--18},
  year={2024},
  organization={Springer}
}

@article{mmJoints,
  title={mmJoints: Expanding Joint Representations Beyond (x, y, z) in mmWave-Based 3D Pose Estimation},
  author={Wang, Zhenyu and Monjur, Mahathir and Nirjon, Shahriar},
  journal={arXiv preprint arXiv:2510.08970},
  year={2025}
}

@article{maepose,
  title={MAEPose: Self-Supervised Spatiotemporal Learning for Human Pose Estimation on mmWave Video},
  author={Wei, Xijia and Fang, Yuan and Chetty, Kevin and Cho, Youngjun and Bianchi-Berthouze, Nadia},
  journal={arXiv preprint arXiv:2605.00242},
  year={2026}
}

@article{gu2024millimeter,
  title={Millimeter wave radar-based human activity recognition for healthcare monitoring robot},
  author={Gu, Zhanzhong and He, Xiangjian and Fang, Gengfa and Xu, Chengpei and Xia, Feng and Jia, Wenjing},
  journal={arXiv preprint arXiv:2405.01882},
  year={2024}
}

@article{hu2024mmwave,
  title={mmWave radar for sit-to-stand analysis: A comparative study with wearables and Kinect},
  author={Hu, Shuting and Ackun, Peggy and Zhang, Xiang and Cao, Siyang and Barton, Jennifer and Hector, Melvin G and Fain, Mindy J and Toosizadeh, Nima},
  journal={IEEE Transactions on Biomedical Engineering},
  volume={72},
  number={9},
  pages={2623--2634},
  year={2025},
  publisher={IEEE}
}

@inproceedings{kong2026twr,
  title={TWR-Pose3D: Through-Wall Radar Based 3D Human Pose Estimation With Dual-Stage Temporal and View Fusion},
  author={Kong, Ailing and Gu, Hengyu and Yan, Xueya and Guo, Jian},
  booktitle={2026 38th Chinese Control and Decision Conference (CCDC)},
  pages={4617--4622},
  year={2026},
  organization={IEEE}
}

@article{pham2026twostage,
  title={A Two-Stage Motion-Aware Framework for mmWave-based Human Mesh Recovery},
  author={Hai Pham, Hoang and Zheng, Shuntian and Li, Jiaqi and Guan, Yu},
  journal={arXiv e-prints},
  pages={arXiv--2605},
  year={2026}
}

@article{hu2026waveman,
  title={WaveMan: mmWave-Based Room-Scale Human Interaction Perception for Humanoid Robots},
  author={Hu, Yuxuan and Zuo, Kuangji and Ma, Boyu and Li, Shihao and Xia, Zhaoyang and Xu, Feng and Yang, Jianfei},
  journal={arXiv preprint arXiv:2601.07454},
  year={2026}
}

@inproceedings{m4human,
  title={M4human: A large-scale multimodal mmwave radar benchmark for human mesh reconstruction},
  author={Fan, Junqiao and Zhou, Yunjiao and Yang, Yizhuo and Cui, Xinyuan and Zhang, Jiarui and Xie, Lihua and Yang, Jianfei and Lu, Chris Xiaoxuan and Ding, Fangqiang},
  booktitle={Proceedings of the IEEE/CVF Conference on Computer Vision and Pattern Recognition},
  pages={42836--42846},
  year={2026}
}

@inproceedings{su2026mmwaveflow,
  title={mmWaveFlow: Unified Enhancement and Generation of mmWave Human Point Clouds},
  author={Su, Chang and Jin, Beihong and Shi, Qiwen and Wang, Zhi},
  booktitle={Proceedings of the IEEE/CVF Conference on Computer Vision and Pattern Recognition},
  pages={31366--31376},
  year={2026}
}

@inproceedings{mahmood2019amass,
  title={AMASS: Archive of motion capture as surface shapes},
  author={Mahmood, Naureen and Ghorbani, Nima and Troje, Nikolaus F and Pons-Moll, Gerard and Black, Michael J},
  booktitle={Proceedings of the IEEE/CVF international conference on computer vision},
  pages={5442--5451},
  year={2019}
}

@incollection{smpl,
  title={SMPL: A skinned multi-person linear model},
  author={Loper, Matthew and Mahmood, Naureen and Romero, Javier and Pons-Moll, Gerard and Black, Michael J},
  booktitle={Seminal Graphics Papers: Pushing the Boundaries, Volume 2},
  pages={851--866},
  year={2023}
}

@inproceedings{pointbert,
  title={Point-bert: Pre-training 3d point cloud transformers with masked point modeling},
  author={Yu, Xumin and Tang, Lulu and Rao, Yongming and Huang, Tiejun and Zhou, Jie and Lu, Jiwen},
  booktitle={Proceedings of the IEEE/CVF conference on computer vision and pattern recognition},
  pages={19313--19322},
  year={2022}
}

@article{pointmae,
  title={Masked autoencoders for 3d point cloud self-supervised learning},
  author={Pang, Yatian and Tay, Eng Hock Francis and Yuan, Li and Chen, Zhenghua},
  journal={World Scientific Annual Review of Artificial Intelligence},
  volume={1},
  pages={2440001},
  year={2023},
  publisher={World Scientific}
}

@inproceedings{maskpoint,
  title={Masked discrimination for self-supervised learning on point clouds},
  author={Liu, Haotian and Cai, Mu and Lee, Yong Jae},
  booktitle={European Conference on Computer Vision},
  pages={657--675},
  year={2022},
  organization={Springer}
}

@article{vqvae,
  title={Neural discrete representation learning},
  author={Van Den Oord, Aaron and Vinyals, Oriol and others},
  journal={Advances in neural information processing systems},
  volume={30},
  year={2017}
}

@inproceedings{duan2022codebook,
  title={Multi-modal alignment using representation codebook},
  author={Duan, Jiali and Chen, Liqun and Tran, Son and Yang, Jinyu and Xu, Yi and Zeng, Belinda and Chilimbi, Trishul},
  booktitle={Proceedings of the IEEE/CVF Conference on Computer Vision and Pattern Recognition},
  pages={15651--15660},
  year={2022}
}

@inproceedings{lee2018scan,
  title={Stacked cross attention for image-text matching},
  author={Lee, Kuang-Huei and Chen, Xi and Hua, Gang and Hu, Houdong and He, Xiaodong},
  booktitle={Proceedings of the European conference on computer vision (ECCV)},
  pages={201--216},
  year={2018}
}

@inproceedings{pct,
  title={Human pose as compositional tokens},
  author={Geng, Zigang and Wang, Chunyu and Wei, Yixuan and Liu, Ze and Li, Houqiang and Hu, Han},
  booktitle={Proceedings of the IEEE/CVF Conference on Computer Vision and Pattern Recognition},
  pages={660--671},
  year={2023}
}

@inproceedings{tokenhmr,
  title={Tokenhmr: Advancing human mesh recovery with a tokenized pose representation},
  author={Dwivedi, Sai Kumar and Sun, Yu and Patel, Priyanka and Feng, Yao and Black, Michael J},
  booktitle={Proceedings of the IEEE/CVF conference on computer vision and pattern recognition},
  pages={1323--1333},
  year={2024}
}

@article{flamingo,
  title={Flamingo: a visual language model for few-shot learning},
  author={Alayrac, Jean-Baptiste and Donahue, Jeff and Luc, Pauline and Miech, Antoine and Barr, Iain and Hasson, Yana and Lenc, Karel and Mensch, Arthur and Millican, Katherine and Reynolds, Malcolm and others},
  journal={Advances in neural information processing systems},
  volume={35},
  pages={23716--23736},
  year={2022}
}

@inproceedings{blip2,
  title={Blip-2: Bootstrapping language-image pre-training with frozen image encoders and large language models},
  author={Li, Junnan and Li, Dongxu and Savarese, Silvio and Hoi, Steven},
  booktitle={International conference on machine learning},
  pages={19730--19742},
  year={2023},
  organization={PMLR}
}

@article{llava,
  title={Visual instruction tuning},
  author={Liu, Haotian and Li, Chunyuan and Wu, Qingyang and Lee, Yong Jae},
  journal={Advances in neural information processing systems},
  volume={36},
  pages={34892--34916},
  year={2023}
}

@article{mmpose,
  title={mm-Pose: Real-time human skeletal posture estimation using mmWave radars and CNNs},
  author={Sengupta, Arindam and Jin, Feng and Zhang, Renyuan and Cao, Siyang},
  journal={IEEE sensors journal},
  volume={20},
  number={17},
  pages={10032--10044},
  year={2020},
  publisher={IEEE}
}

@inproceedings{mmmesh,
  title={mmMesh: Towards 3D real-time dynamic human mesh construction using millimeter-wave},
  author={Xue, Hongfei and Ju, Yan and Miao, Chenglin and Wang, Yijiang and Wang, Shiyang and Zhang, Aidong and Su, Lu},
  booktitle={Proceedings of the 19th annual international conference on mobile systems, applications, and services},
  pages={269--282},
  year={2021}
}

@inproceedings{m4esh,
  title={M4esh: mmwave-based 3d human mesh construction for multiple subjects},
  author={Xue, Hongfei and Cao, Qiming and Ju, Yan and Hu, Haochen and Wang, Haoyu and Zhang, Aidong and Su, Lu},
  booktitle={Proceedings of the 20th ACM Conference on Embedded Networked Sensor Systems},
  pages={391--406},
  year={2022}
}

@inproceedings{p4transformer,
  title={Point 4d transformer networks for spatio-temporal modeling in point cloud videos},
  author={Fan, Hehe and Yang, Yi and Kankanhalli, Mohan},
  booktitle={Proceedings of the IEEE/CVF conference on computer vision and pattern recognition},
  pages={14204--14213},
  year={2021}
}

@article{mri,
  title={mri: Multi-modal 3d human pose estimation dataset using mmwave, rgb-d, and inertial sensors},
  author={An, Sizhe and Li, Yin and Ogras, Umit},
  journal={Advances in neural information processing systems},
  volume={35},
  pages={27414--27426},
  year={2022}
}

@inproceedings{hupr,
  title={Hupr: A benchmark for human pose estimation using millimeter wave radar},
  author={Lee, Shih-Po and Kini, Niraj Prakash and Peng, Wen-Hsiao and Ma, Ching-Wen and Hwang, Jenq-Neng},
  booktitle={Proceedings of the IEEE/CVF Winter Conference on Applications of Computer Vision},
  pages={5715--5724},
  year={2023}
}

@inproceedings{mmbody,
  title={mmbody benchmark: 3d body reconstruction dataset and analysis for millimeter wave radar},
  author={Chen, Anjun and Wang, Xiangyu and Zhu, Shaohao and Li, Yanxu and Chen, Jiming and Ye, Qi},
  booktitle={Proceedings of the 30th ACM International Conference on Multimedia},
  pages={3501--3510},
  year={2022}
}

@article{mmfi,
  title={Mm-fi: Multi-modal non-intrusive 4d human dataset for versatile wireless sensing},
  author={Yang, Jianfei and Huang, He and Zhou, Yunjiao and Chen, Xinyan and Xu, Yuecong and Yuan, Shenghai and Zou, Han and Lu, Chris Xiaoxuan and Xie, Lihua},
  journal={Advances in Neural Information Processing Systems},
  volume={36},
  pages={18756--18768},
  year={2023}
}

@inproceedings{rtpose,
  title={Rt-pose: A 4d radar tensor-based 3d human pose estimation and localization benchmark},
  author={Ho, Yuan-Hao and Cheng, Jen-Hao and Kuan, Sheng Yao and Jiang, Zhongyu and Chai, Wenhao and Huang, Hsiang-Wei and Lin, Chih-Lung and Hwang, Jenq-Neng},
  booktitle={European Conference on Computer Vision},
  pages={107--125},
  year={2024},
  organization={Springer}
}

@article{immfusion,
  title={Immfusion: Robust mmwave-rgb fusion for 3d human body reconstruction in all weather conditions},
  author={Chen, Anjun and Wang, Xiangyu and Shi, Kun and Zhu, Shaohao and Fang, Bin and Chen, Yingfeng and Chen, Jiming and Huo, Yuchi and Ye, Qi},
  journal={arXiv preprint arXiv:2210.01346},
  year={2022}
}

@inproceedings{emdul,
  title={Expanding mmWave Datasets for Human Pose Estimation with Unlabeled Data and LiDAR Datasets},
  author={Peng, Zhuoxuan and Zhu, Boan and Zhang, Xingjian and Li, Wenying and Chan, S-H Gary},
  booktitle={Proceedings of the IEEE/CVF Conference on Computer Vision and Pattern Recognition},
  pages={21221--21230},
  year={2026}
}

@article{wicompass,
  title={WiCompass: Oracle-driven Data Scaling for mmWave Human Pose Estimation},
  author={Liang, Bo and Gong, Chen and Wang, Haobo and Liu, Qirui and Zhou, Rungui and Shao, Fengzhi and Wang, Yubo and Gao, Wei and Zhou, Kaichen and Cui, Guolong and others},
  journal={arXiv preprint arXiv:2602.18726},
  year={2026}
}

@article{dghmesh,
  title={DGHMesh: A Large-scale Dual-radar mmWave Dataset and Generalization-focused Benchmark for Human Mesh Reconstruction},
  author={Guo, Rongxiao and Chen, Qingchao},
  journal={arXiv preprint arXiv:2604.22827},
  year={2026}
}

@inproceedings{m3track,
  title={m3track: mmwave-based multi-user 3d posture tracking},
  author={Kong, Hao and Xu, Xiangyu and Yu, Jiadi and Chen, Qilin and Ma, Chenguang and Chen, Yingying and Chen, Yi-Chao and Kong, Linghe},
  booktitle={Proceedings of the 20th Annual International Conference on Mobile Systems, Applications and Services},
  pages={491--503},
  year={2022}
}

@inproceedings{mmpoint,
  title={mmPoint: Dense Human Point Cloud Generation from mmWave.},
  author={Xie, Qian and Deng, Qianyi and Cheng, Ta Ying and Zhao, Peijun and Patel, Amir and Trigoni, Niki and Markham, Andrew},
  booktitle={BMVC},
  pages={194--196},
  year={2023}
}

@inproceedings{transhupr,
  title={TransHuPR: Cross-View Fusion Transformer for Human Pose Estimation Using mmWave Radar.},
  author={Kini, Niraj Prakash and Shiue, Ruey-Horng and Chandra, Ryan and Peng, Wen-Hsiao and Ma, Ching-Wen and Hwang, Jenq-Neng},
  booktitle={BMVC},
  year={2024}
}

@inproceedings{vqhps,
  title={Vq-hps: Human pose and shape estimation in a vector-quantized latent space},
  author={Fiche, Gu{\'e}nol{\'e} and Leglaive, Simon and Alameda-Pineda, Xavier and Agudo, Antonio and Moreno-Noguer, Francesc},
  booktitle={European Conference on Computer Vision},
  pages={471--490},
  year={2024},
  organization={Springer}
}

@article{cpformer,
  title={CPFormer: End-to-end multi-person human pose estimation from raw radar cubes with transformers},
  author={Chen, Lin and Wang, Guoli},
  journal={IEEE Sensors Journal},
  year={2025},
  publisher={IEEE}
}

@article{mmhpe,
  title={mmhpe: Robust multiscale 3-d human pose estimation using a single mmwave radar},
  author={Wu, Yingxiao and Jiang, Zhongmin and Ni, Haocheng and Mao, Changlin and Zhou, Zhiyuan and Wang, Wenxiang and Han, Jianping},
  journal={IEEE Internet of Things Journal},
  volume={12},
  number={1},
  pages={1032--1046},
  year={2024},
  publisher={IEEE}
}

@inproceedings{mvdopplerpose,
  title={MVDoppler-Pose: Multi-modal multi-view mmWave sensing for long-distance self-occluded human walking pose estimation},
  author={Choi, Jaeho and Hor, Soheil and Yang, Shubo and Arbabian, Amin},
  booktitle={Proceedings of the Computer Vision and Pattern Recognition Conference},
  pages={27750--27759},
  year={2025}
}

@inproceedings{dptm,
  title={Dilated Point Spatio-Temporal Mesh Transformer for mmWave Radar-Based Human Mesh Reconstruction},
  author={Chen, Lin and Wang, Guoli},
  booktitle={2025 International Joint Conference on Neural Networks (IJCNN)},
  pages={1--8},
  year={2025},
  organization={IEEE}
}

@article{posegraphnet,
  title={PoseGraphNet: Pose prior and graph structure for 3D human pose estimation using mmWave radar},
  author={Su, Yuanzhi and Hou, Huiying Cynthia and Zhao, Chun},
  journal={Measurement},
  pages={118851},
  year={2025},
  publisher={Elsevier}
}

@inproceedings{millimamba,
  title={milliMamba: Specular-Aware Human Pose Estimation via Dual mmWave Radar with Multi-Frame Mamba Fusion},
  author={Kini, Niraj Prakash and Tsai, Shiau-Rung and Lin, Guan-Hsun and Peng, Wen-Hsiao and Ma, Ching-Wen and Hwang, Jenq-Neng},
  booktitle={Proceedings of the IEEE/CVF Winter Conference on Applications of Computer Vision},
  pages={1481--1490},
  year={2026}
}

@article{umimo,
  title={UMIMO: Universal unsupervised learning for mmWave radar sensing with MIMO array synthesis},
  author={Zhang, Haoyu and Zhang, Dongheng and Song, Ruiyuan and Wu, Zhi and Chen, Jinbo and Fang, Liang and Lu, Zhi and Hu, Yang and Lin, Hui and Chen, Yan},
  journal={IEEE Transactions on Mobile Computing},
  year={2025},
  publisher={IEEE}
}

@article{mmrehab,
  title={Sensing Life in Stillness: Unified Dynamic and Static Human Mesh Reconstruction with mmWave Radar},
  author={Chen, Lin and Li, Cong and Zhong, Shuxin and Chen, Jun and Wen, Yufei and Song, Haotian and Wu, Kaishun},
  journal={Proceedings of the ACM on Interactive, Mobile, Wearable and Ubiquitous Technologies},
  volume={10},
  number={1},
  pages={1--25},
  year={2026},
  publisher={ACM New York, NY, USA}
}

@inproceedings{syncheck,
  title={Data Can Speak for Itself: Quality-guided Utilization of Wireless Synthetic Data},
  author={Gong, Chen and Liang, Bo and Gao, Wei and Xu, Chenren},
  booktitle={Proceedings of the 23rd Annual International Conference on Mobile Systems, Applications and Services},
  pages={209--222},
  year={2025}
}

@inproceedings{wiswiss,
  title={Towards Generalizable Wireless Sensing Models via Pre-training on Multi-Source Datasets},
  author={Gong, Chen and Liang, Bo and Zhu, Qihao and Gao, Wei and Chen, Yin and Nakazawa, Jin and Xu, Chenren},
  booktitle={Proceedings of the 2026 ACM/IEEE International Conference on Embedded Artificial Intelligence and Sensing Systems},
  pages={488--502},
  year={2026}
}

@inproceedings{zhang2026mupose,
  title={MuPose: Breaking the Scalability Barrier of mmWave Multi-User Pose Estimation in the Wild},
  author={Zhang, Duo and Yin, Zhehui and Zhang, Xusheng and Wang, Junzhe and Yang, Hongliu and Yao, Zhiyun and Fan, Zizhou and Li, Wenwei and Zhang, Daqing},
  booktitle={Proceedings of the 24th Annual International Conference on Mobile Systems, Applications and Services},
  pages={622--636},
  year={2026}
}

@article{yang2026bridging,
  title={Bridging the Resolution Gap: Cost-Effective Human Point Cloud Generation via Low-Bandwidth mmWave Radar},
  author={Yang, Hongliu and Fan, Zizhou and Wang, Yueyang and Xiong, Jie and Zhang, Duo and Zhang, Xusheng and Wang, Junzhe and Zhang, Fusang and Zhang, Daqing},
  journal={Proceedings of the ACM on Interactive, Mobile, Wearable and Ubiquitous Technologies},
  volume={10},
  number={2},
  pages={1--26},
  year={2026},
  publisher={ACM New York, NY, USA}
}

@article{zhang2023lt,
  title={Lt-fall: The design and implementation of a life-threatening fall detection and alarming system},
  author={Zhang, Duo and Zhang, Xusheng and Li, Shengjie and Xie, Yaxiong and Li, Yang and Wang, Xuanzhi and Zhang, Daqing},
  journal={Proceedings of the ACM on Interactive, Mobile, Wearable and Ubiquitous Technologies},
  volume={7},
  number={1},
  pages={1--24},
  year={2023},
  publisher={ACM New York, NY, USA}
}
}

\end{document}